*Article*

# Multi-Finger Haptics: Analysis of Human Hand Grasp towards a Tripod Three-Finger Haptic Grasp model

Jose James

Brown University, Providence, RI, 02912, USA.
Correspondence: jose_james@brown.edu.

**Abstract:** Grasping is an incredible ability of animals using their arms and limbs in their daily life. The human hand is an especially astonishing multi-fingered tool for precise grasping, which helped humans to develop the modern world. The implementation of the human grasp to virtual reality and tele robotics is always interesting and challenging at the same time. In this work, authors surveyed, studied, and analyzed the human hand grasping behavior for the possibilities of haptic grasping in the virtual and remote environment. This work is focused on the motion and force analysis of fingers in human hand grasping scenarios and the paper describes the transition of the human hand grasping towards a tripod haptic grasp model for effective interaction in virtual reality.

## 1. Introduction

The human hand is a highly skilled, prehensile, multi-fingered, perception and manipulation organ at the distal end of the arm [1]. Prehensility is the quality of an appendage or organ that has adapted for grasping or holding. In the past decade's, researchers [2] explored the various aspects of the evolution, morphology, anthropology, and social significance of the human hand as a tool [3], as a symbol [4] and as a weapon [5]. Humans cognitively manipulating a variety of objects in daily life using various hand configurations, resulting from changing the position, orientation and placement of hand and fingers based on the object properties such as its weight, shape, texture, friction, hardness etc. Such a variety of grasps is possible because of the dexterity, various degrees of freedom, and the great control strategy.

Hands are associated with the aligning capability of the body, kinesthetic perception of the limb and the richest tactile sense. Many researchers studied the anatomy, muscles, biomechanics, kinematics, functionalities, and skills of human hand [6,7]. These studies helped to do more focused research on human hand prehension [8] and grasp [9,10]. Based on all these studies hand grasp types are classified and different grasp taxonomies were arising in the literature [11,12], covering a broad range of domains.

As technologies like virtual reality and tele-robotics being progressed, humans started interacting with virtual and remote environments. But the integration of hand grasping into virtual and remote environments still challenging because of the complex architecture behind it. The extensive research in the analysis and synthesis of human grasps [13,14] over the past years has provided a basic theoretical framework towards better progress in human-computer interaction [15], robotic grasping [16] and dexterous manipulation and lead to the design of artificial robotics hands and arms for the prosthetic application [17]. Since the last few decades, researchers put effort to mimic the human hand to design robotic grippers [18], etc. But still, these frameworks need more extensive studies for the practical implementation of the direct involvement of humans to grasp objects in virtual and remote environments with multi-fingers.

The grasping force depends on the orientation of fingers, palm, and wrist [19]. The force output on the fingertip is highly joint dependent and provides stable grasp and precise manipulation of objects. Previous works [20], more focused on muscle activation patterns and resultant positions/forces as a function of the joints as well as subject independent leads to the structural variability in human hands [21]. Motion and force analysis of fingers in various hand manipulation actions can be observed, learned, and analyzed to come up with better framework and devices for virtual and remote manipulations [22,23] Adapting the gripping functions, manipulation capabilities, kinematics, dynamics, and size of the human hand, will accelerate the design of the human-like artificial arms and hands for the direct grasping interaction in the virtual world.

As a previous work, authors designed haptic interfaces for tweezer pinch grasp [24] and tripod grasp [25]. Also implemented the hand grasp through augmented haptics by means of custom-made attachments for virtual tools in motor skill training interfaces [26,27]. The aim of this work is to understand human grasping and manipulations, surveying different types of grasp taxonomies, study the characterization of hand in grasping, model a tripod haptic grasp and design an interface for multi-finger haptic grasping which can offer better interactions with tasks in the virtual and remote environments.

## 2. Human Hand

The human hand is a prehensile, multi-fingered astonishing organ/tool of complex engineering used to carry and manipulate objects [2]. In view of human grasping, a short description of hand anatomy, mechanisms and kinematics will help to model a multi-finger haptic grasp and design a haptic grasping interface.

### 2.1. Hand Anatomy

The human hand includes mainly three areas and five digits (fingers): Thumb, Index finger, Middle finger, Ring finger, and Little finger are numbered 1-5 as shown in Figure 1. The palm with fingers holds most pressure and support for the hand to grasp. Fingers are the densest areas of nerve endings and the richest source of tactile feedback. So, hands are the primary tool for a sense of touch and positioning capability. The human hand consists of 27 bones and 45 muscles with at least 23 degrees of freedom at the joints [6] including the wrist as shown in Figure 1(a).

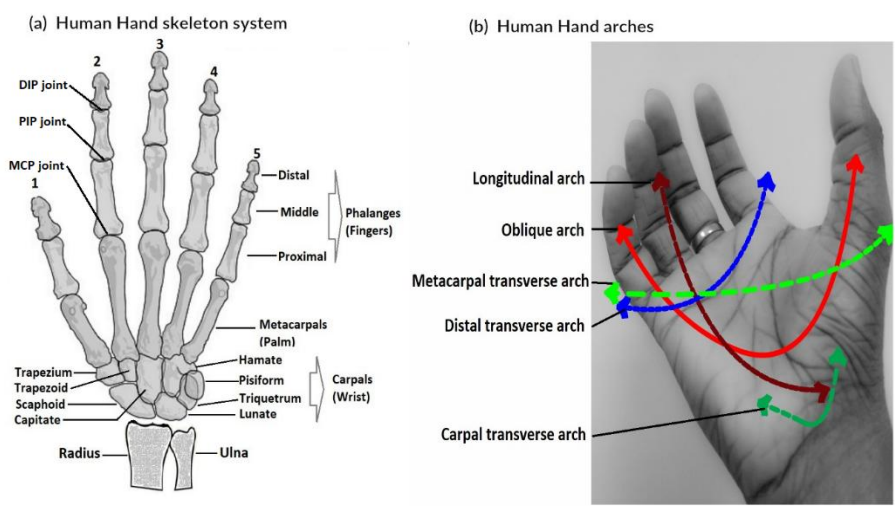

**Figure 1.** Skeleton system and arches of human hand

The human hand can grasp objects and do daily tasks by forming bony arches: Longitudinal arches, transverse arches, and oblique arches as drawn in Figure 1(b). Lon-gitudinal arches shaped by the finger bones and their associated metacarpal bones, transverse arches by the carpal bones and the distal ends of the metacarpal bones, and oblique arches by the thumb and four fingers [7]. These arches are the basic frames for various grasp patterns. The extrinsic and intrinsic muscles in hand controlled the motion of the fingers and making grasping possible. Thumb together with the index and middle finger forms the dynamic tridactyl configuration responsible for most grips which not requiring force. The ring and little fingers are more static. So, for this multi-finger haptic grasping interface model, authors considered the motion of the first three fingers the thumb, index finger and middle finger.

*2.2 Hand Kinematics*

Hand's numerous patterns of action was resulted from the skeleton mechanical systems and the twenty-four muscle groups regulated by the diverse motor and sensory nerve pathways [6].

The MCP and PIP joints exhibit a common rotation pattern. The virtual center of rotation of hand is the center of curvature of the distal end of the proximal member [28]. The lateral rotation of fingers is small in the MCP joints and decreasing towards the phalangeal hinge joints. The thumb has the greater mobility in the CMC articulation. Other fingers being more arched from index to little finger. The thumb, palm, and fingers together permitted to grasp a 1.75-inch cylinder at about 45 degrees to the radioulnar axis. Bunnell [29] considers this "an ancestral position ready for grasping limbs, weapons, or other creatures."

The major wrist motions are extension (or dorsiflexion), flexion (or volar flexion), radial flexion and ulnar flexion, based on the angle of rotation of the wrist. The fixation movements and ballistic movements are also major types of movements in the hand [30]. The hand with the fully extended arm can be rotated through almost 360 degrees with the participation of shoulder and elbow. From palm up to palm down, the hand can be rotated through 180 degrees, with the elbow flexed. Thumb can provide a variety of flexions extension patterns of the phalanges for any given metacarpal position and due to the relative mobility of the CMC joint, which allows the thumb to act in any plane necessary to oppose the digits. In the principal opposition cases and prehensions, the plane of the thumb action is inclined 45 to 60 degrees to the palmar plane. In lateral prehension, the plane is approximately parallel to the palmar plane.

*2.3 Hand Dynamics*

Fick [31] investigated the actions and contractile forces of hand muscles and estimated the summed forces of the individual muscles participating in the action. But the measured isometric forces are only 10% of the total forces because of the effective small moment arm upon any of the wrist or hand joints. The flexor-extensor forces in the wrist and the prehensile forces in the hand varied with wrist angle and it reaches a maximum at a wrist angle of about 145 degrees. So, for very strong prehensions, wrist likely to attain this angle [32].

Kamper et.al [33] was analyzed the joint angles and finger trajectories in reach-and-grasp tasks which fit the actual finger positions with a mean error of 0.23 ± 0.25 cm and accounted for over 98% of the variance in finger position. The direction of the thumb trajectories exhibited a greater dependence on object type than the finger trajectories, but still utilized a small percentage (<5%) of the available workspace [34]. Previous studies of musculoskeletal models [35] observed that the role of the intrinsic muscles of the hand as the main force-producing muscles in power grip [36]. These models were used in the commercial software ANYBODY for the thumb and the index finger [37]. However, these models did not address the coupling between the fingers and the interaction with the wrist which limits the investigation of human grasping.

In [38] authors presented an upper limb musculature model for the full arm including the shoulder, elbow, wrist, thumb, and index finger and provides valuable data on the wrist-finger joint coupling and extrinsic hand muscle anatomy [39]. The force output on the fingertip is highly joint dependent and provides stable grasp and precise manipulation of objects. That's why the force transfer is important for the human hand in contact and manipulating purpose to avoid the slipping and deformation of the object. Previous works focused more on muscle activation patterns and resultant positions/forces as a function of the joints [20] as well as a subject independent lead to the structural variability in human hands.

*2.4 Purpose of Study*

This study conducted to analyze the human hand grasping to isolate the functional properties with the final goal to optimize haptic grasping by building simpler multi-finger haptic grasping interfaces with at least similar grasping and manipulation capabilities. This work helped to learn more about the complex engineering structure of the human hand and leads to characterizing the multi-finger haptic grasping systems. For instance, the proposed three-finger haptic grasping system has an independent joint architecture for three fingers in the hand, which may be advantageous in virtual grasping. Also, this led to identify the independence of joints needs in normal grasping tasks. Conversely, the control of such an independent architecture is very challenging. By adding synergies, we can reduce the complexity of control, but we also want to keep a certain, currently unquantified, level of dexterity.

**3. Human Grasp**

A grasp is a system wherein the desired object is gripped by the fingers of a human (or robot) hand.

*3.1 Human Grasp Patterns*

Napier [40] categorized human grasp into two basic grips: power grasps and precision grasps (pinch grasps). In power grasp, the object is in the palm of the hand and enclosed by the fingers which lead to large area of contact between the palm, the fingers, and the object. In precision grasp (pinch grip), the object is held between the tip of the thumb and finger, which offer more dexterity. Precision grasps become more relevant in robotic and virtual grasping. The power grasp is enhanced by the precision grasp between the thumb and the distal finger pads, and it is inherently stable. Pinch grip requires the six joints between the index finger and the thumb to be stabilized; it requires more activity of the intrinsic finger muscles to maintain this balance. A large variety of prehension patterns are identified from studies of the muscle-bone-joint anatomy and from observation of the postures and motions of the hand. The object-contact pattern furnishes a satisfactory basis for classification of major prehension patterns [41].

All the ages, the human hand was a part of most creative arts of every culture [42] to speak and convey human emotions and the hands symbolize cultural behaviors, values, and beliefs. A mudra is a symbolic gesture in the spiritual practice of Indian religions and traditional art forms performed with the hands with a specific pattern of finger configurations [43]. A canonical set of predefined hand postures and modifiers can be used in digital human modelling to develop the standard hand posture libraries and a universal referencing scheme and continuum of hand poses from simple posture to complex one. Researchers [9] have studied features for force-closure grasp by human hands and characterized into four mutually independent properties for robotic arm grasping listed as dexterity, equilibrium, stability, and dynamic behavior. The principal component analysis of static hand posture of several subjects provides information about the finger joint variance and shape of the grasped object and did not consider the hand position/orientation relative to the object placement [44].

*3.2 Grasp Taxonomy*

Based on the various studies about human hand in the literature, hand grasp types are classified, and different grasp taxonomies were raised in the literature [11,12] as shown in Table 1. If the grasp object size/shape is not considered, this taxonomy might be lowered to broad range.

Grasps are classified based on precision [45], grasped object's size [9], shape [46], weight, rigidity, force requirement, the position of thumb (adducted or abducted position) and the situation. Based on the level of precision, grasps are classified as precision grasp [10], intermediate grasp [47], and power grasp [48]. The movements of the hand in the power grip are evoked by the arm but in the precision handling, the intrinsic movements on the hand not evoked by the arm. In intermediate grasp, elements of power and precision grasps are present in roughly the same proportion. Later studies included other grasps like hook grasp, flat hand grasp, platform grasp, push grasp [49]. In static and stable grasps [12], the object is in a constant relation to the hand.

Based on the direction of force relative to the hand coordinate frame, applied by the hand on the object to hold it securely [8] opposition type grasps are classified as pad opposition, palm opposition and side opposition grasp [50]. Pad Opposition occurs between hand surfaces along a direction generally parallel to the palm, usually occurs between volar (palmar) surfaces and the fingers and thumb, near or on the pads. Palm Opposition occurs between hand surfaces along a direction generally perpendicular to the palm. Side Opposition occurs between hand surfaces along a direction generally transverse to the palm.

The taxonomy of Cutkosky [9], which is widely used in the field of robotics, lists 15 different grasps. Other taxonomies mentioned in works of Kamakura et al. [47], Edwards et al. [10], Kapandji [45] are listed with 14, 20 and 21 grasps respectively. A similar study [51], that used a different categorization which incorporated non-prehensile grasps. Even though there has been a considerable effort in creating statistics of human hand use and grouping of hand grasps [11,12]. The extensive research in human grasp analysis and taxonomies over the past years helped towards better progress in human-computer interaction, robotic grasping, and dexterous manipulation and lead to the design of artificial robotics hands and arms for the prosthetic application.

This helped to classify all hand usage in everyday life situations [52]. Furthermore, the taxonomy could be extended to include non-prehensile "grasps", or for dynamic within - hand manipulation movements. Kamakura et al. [47] classified the tripod grasps as intermediate grasps, apart from that it was classified as a precision grasp. Several studies have investigated classifying grasps into a discrete set of types [8,9,47], and others have been aimed at understanding certain aspects of human hand usage [53]. The number of fingers used for grasping increases with the size and mass of the object [54] until a two-handed grasp is required, indicating that object size and mass are strong factors in determining the grasp type. The rigidity of the objects also influencing the grasp [55].

| No. | Grasp Name | Picture | Type | Opp.Type | Thumb Position | Virtual Fingers | Mod. VF | Min. No. of Fingers | No. | Grasp Name | Picture | Type | Opp.Type | Thumb Position | Virtual Fingers | Mod. VF | Min. No. of Fingers |
|---|---|---|---|---|---|---|---|---|---|---|---|---|---|---|---|---|---|
| 1 | large Diameter | 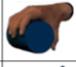 | Power | Palm | Abducted | VF1:<br>VF2: 2-5<br>VF3: | VF1: 1<br>VF2: 2-5<br>VF3: Palm | 3 | 18 | Extension Type | 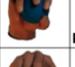 | Power | Pad | Abducted | VF1:<br>VF2: 2-5<br>VF3: | VF1: 1<br>VF2: 2-5<br>VF3: | 3 |
| 2 | Small Daimeter | 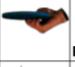 | Power | Palm | Abducted | VF1:<br>VF2: 2-5<br>VF3: | VF1: Palm<br>VF2: 2-5<br>VF3: | 2 | 19 | Distal Type | 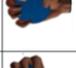 | Power | Pad | Abducted | VF1:<br>VF2: 2-5<br>VF3: | VF1: 1<br>VF2: 2-5<br>VF3: | 2 |
| 3 | Medium Wrap. | 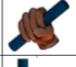 | Power | Palm | Abducted | VF1:<br>VF2: 2-5<br>VF3: | VF1: Palm<br>VF2: 2-5<br>VF3: | 3 | 20 | Writing Tripod | 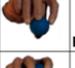 | Precision | Side | Abducted | VF1:<br>VF2: 2<br>VF3: | VF1: 1<br>VF2: 2<br>VF3: 3 | 3 |
| 4 | Adducted Thumb | 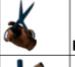 | Power | Palm | Adducted | VF1:<br>VF2: 2-5<br>VF3: 1 | VF1: Palm<br>VF2: 2-5<br>VF3: 1 | 3 | 21 | Tripod Variation | 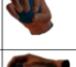 | Intermediate | Side | Abducted | VF1:<br>VF2: 3-4<br>VF3: | VF1: 1<br>VF2: 3-4<br>VF3: 2 | 3 |
| 5 | Light Tool | 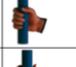 | Power | Palm | Adducted | VF1:<br>VF2: 2-5<br>VF3: (1) | VF1: Palm<br>VF2: 2-5<br>VF3: (1) | 3 | 22 | Parallel Extension | 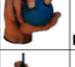 | Precision | Pad | Adducted | VF1:<br>VF2: 2-5<br>VF3: | VF1: 1<br>VF2: 2-5<br>VF3: | 3 |
| 6 | Prismatic 4 Finger | 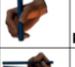 | Precision | Pad | Abducted | VF1:<br>VF2: 2-5<br>VF3: | VF1: 1<br>VF2: 2<br>VF3: 3-5 | 3 | 23 | Adduction Grip | 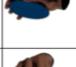 | Intermediate | Side | Abducted | VF1:<br>VF2: 2<br>VF3: | VF1: 2<br>VF2: 3<br>VF3: | 2 |
| 7 | Prismatic 3 Finger | 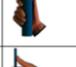 | Precision | Pad | Abducted | VF1:<br>VF2: 2-4<br>VF3: | VF1: 1<br>VF2: 2<br>VF3: 3-4 | 3 | 24 | Tip Pinch | 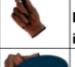 | Precision | Pad | Abducted | VF1:<br>VF2: 2<br>VF3: | VF1: 1<br>VF2: 2<br>VF3: | 2 |
| 8 | Prismatic 2 Finger | 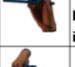 | Precision | Pad | Abducted | VF1:<br>VF2: 2-3<br>VF3: | VF1: 1<br>VF2: 2<br>VF3: 3 | 3 | 25 | Lateral Tripod | 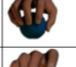 | Intermediate | Side | Adducted | VF1:<br>VF2: 3<br>VF3: | VF1: 1-2<br>VF2: 3<br>VF3: | 3 |
| 9 | Palmar Pinch | 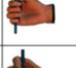 | Precision | Pad | Abducted | VF1:<br>VF2: 2<br>VF3: | VF1: 1<br>VF2: 2<br>VF3: | 2 | 26 | Sphere 4 Finger | 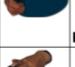 | Power | Pad | Abducted | VF1:<br>VF2: 2-4<br>VF3: | VF1: 1<br>VF2: 2-4<br>VF3: | 3 |
| 10 | Power Disk | 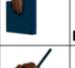 | Power | Palm | Abducted | VF1:<br>VF2: 2-5<br>VF3: | VF1: 1<br>VF2: 2-5<br>VF3: Palm | 3 | 27 | QuadPod | 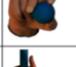 | Precision | Pad | Abducted | VF1:<br>VF2: 2-4<br>VF3: | VF1: 1<br>VF2: 2-4<br>VF3: | 3 |
| 11 | Power Sphere | 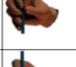 | Power | Palm | Abducted | VF1:<br>VF2: 2-5<br>VF3: | VF1: 1<br>VF2: 2-5<br>VF3: Palm | 3 | 28 | Sphere 3 Finger | 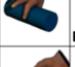 | Power | Pad | Abducted | VF1:<br>VF2: 2-3<br>VF3: | VF1: 1<br>VF2: 2-3<br>VF3: | 3 |
| 12 | Precision Disk | 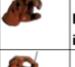 | Precision | Pad | Abducted | VF1:<br>VF2: 2-5<br>VF3: | VF1: 1<br>VF2: 2-5<br>VF3: | 3 | 29 | Stick | 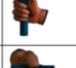 | Intermediate | Side | Adducted | VF1:<br>VF2: 2<br>VF3: | VF1: 1<br>VF2: 2<br>VF3: 3-5 | 3 |
| 13 | Precision Sphere | 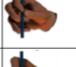 | Precision | Pad | Abducted | VF1:<br>VF2: 2-5<br>VF3: | VF1: 1<br>VF2: 2-5<br>VF3: | 3 | 30 | Palmar | 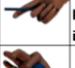 | Power | Palm | Adducted | VF1:<br>VF2: 2-5<br>VF3: | 1,Palm<br>VF2: 2-5<br>VF3: | 3 |
| 14 | Tripod | 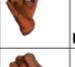 | Precision | Pad | Abducted | VF1:<br>VF2: 2-3<br>VF3: | VF1: 1<br>VF2: 2-3<br>VF3: | 3 | 31 | Ring | 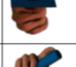 | Power | Pad | Abducted | VF1:<br>VF2: 2<br>VF3: | VF1: 1<br>VF2: 2<br>VF3: | 2 |
| 15 | Fixed Hook | 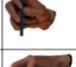 | Power | Palm | Adducted | VF1:<br>VF2: 2-5<br>VF3: | VF1: Palm<br>VF2: 2-5<br>VF3: | 3 | 32 | Ventral | 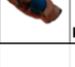 | Intermediate | Side | Adducted | VF1:<br>VF2: 2<br>VF3: | VF1: 1<br>VF2: 2<br>VF3: 3-5 | 3 |
| 16 | Lateral | 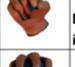 | Intermediate | Side | Adducted | VF1:<br>VF2: 2<br>VF3: | VF1: 1<br>VF2: 2<br>VF3: 3-5 | 2 | 33 | Inferier Pincer | 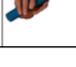 | Precision | Pad | Abducted | VF1:<br>VF2: 2<br>VF3: | VF1: 1<br>VF2: 2<br>VF3: | 2 |
| 17 | Index Finger Extension | 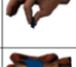 | Power | Palm | Adducted | VF1:<br>VF2: 3-5<br>VF3: 2 | VF1: 1<br>VF2: 3-5<br>VF3: 2 | 3 | | | | | | | | | |

**Table 1.** Taxonomy of grasps and allocation of virtual fingers.

The studies in [56,13,14] analyzed the human grasping scenarios and behavior of unstructured tasks and investigated the relationship between grasp types and object properties and the results indicated that three-fingertip precision grasps such as thumb-2 finger, tripod, or lateral tripod can be used to handle dexterous manipulation of a wide range of objects. In [12] authors analyzed and compared 33 existing human grasp taxonomies of static and stable grasps performed by one hand and synthesized them into a single new taxonomy called 'The GRASP Taxonomy'. The grasps are arranged according to opposition type, the virtual finger assignments, type in terms of power, precision or intermediate grasp, and the position of the thumb. The classifications of micro-interaction grasp instances [57] helped researchers in defining human hand capabilities [58] and affordances in robotic hand design [59].

The concept of the virtual finger [15] has also incorporated in this taxonomy as an abstract representation through which the human brain plans grasping tasks [60]. The virtual finger is a functional unit of several fingers work together comprised of at least one real physical finger to reduce the degrees of the human hand to perform the grasping task. This concept replaces the analysis of the mechanical degrees of freedom of in-dividual fingers by the analysis of the functional roles of forces being applied in a grasp. The virtual fingers oppose each other in the grasp. Virtual fingers are assigned for each grasp in the grasp taxonomy as mentioned in Table 1. Our characterization study revised the existing virtual fingers allocation and replaced with new as mentioned in Table 1. In this work, the authors aim for designing a three-finger haptic grasping interface for virtual grasping the common objects in everyday life and tools in motor skill professions. For the proposed three-finger haptic grasping interface, the thumb assigned as $VF_1$, Index finger as $VF_2$ and other three fingers as a single virtual finger $VF_3$ as explained in section 5.

## 4. Characterization Experiments

*4.1 Overview*

For designing a multi-finger haptic grasping interface, it is quintessential to study the characteristics of the motion and force distribution of fingers in grasping activities. This characterization study helped to propose models for multi-finger haptic rendering and grasping haptic devices. Here the authors conducted experiments for calculating the finger movements, trajectories, positions, orientations on different grasping activities. Also tracked the positions and orientations of the skeleton of each finger through motion tracking techniques. A characterization study was carried out to compute the models and the design of the multi-finger haptic device.

*4.2 Subjects and Methods*

Six subjects, 3 males and 3 females between the ages of 23 and 35 years were taking part in this grasping experiment study with 10 common objects as listed in Table 2 with possible grasp patterns and minimum number of virtual fingers to execute the grasp. Each subject grasps each object for 5 seconds and repeats for 5 trials results total 300 instances for the data sets. This study aimed on the force distribution and orientation of wrist, palm, and finger in all grasping scenarios. The experiment set up consists of a leap motion sensor [61] to track the movement of fingers and a wearable glove with Force Sensitive Resistor (FSR) [62] to measure the force on fingers while grasping different objects and a computer display with the virtual grasping interface as shown in Figure 2.

| Grasp Pattern | Large Diameter | Small Daimeter | Medium Wrap. | Adducted Thumb | Light Tool | Prismatic 4 Finger | Prismatic 3 Finger | Prismatic 2 Finger | Palmar Pinch | Power Disk | Power Sphere | Precision Disk | Precision Sphere | Tripod | Fixed Hook | Lateral | Index Finger Extension | Extension Type | Distal Type | Writing Tripod | Tripod Variation | Parallel Extension | Adduction Grip | Tip Pinch | Lateral Tripod | Sphere 4 Finger | QuadPod | Sphere 3 Finger | Stick | Palmar | Ring | Ventral | Inferier Pincer |
|---|---|---|---|---|---|---|---|---|---|---|---|---|---|---|---|---|---|---|---|---|---|---|---|---|---|---|---|---|---|---|---|---|---|
| min.num. of VF | 3 | 2 | 3 | 3 | 3 | 3 | 3 | 2 | 3 | 3 | 3 | 3 | 3 | 3 | 3 | 2 | 3 | 3 | 2 | 3 | 3 | 3 | 2 | 2 | 3 | 3 | 3 | 3 | 3 | 3 | 2 | 3 | 2 |
| **Objects** | | | | | | | | | | | | | | | | | | | | | | | | | | | | | | | | | |
| Ball Pen | | ✓ | | | ✓ | ✓ | ✓ | ✓ | | | | | | | | | | | | ✓ | | | ✓ | ✓ | | | | | ✓ | | | ✓ | |
| Marker Pen | | ✓ | ✓ | ✓ | | | | | | | | | | ✓ | | | ✓ | | | ✓ | | | | | | | | | ✓ | | | ✓ | |
| Cube box | | | | | | | | | | | | | ✓ | | | | | | | | | | ✓ | | ✓ | ✓ | | ✓ | | | | | ✓ |
| Toy Wheel | | | | | | | | | ✓ | | ✓ | | ✓ | | | | | | | ✓ | | | | | ✓ | | ✓ | | ✓ | | | | ✓ |
| Cup | ✓ | | | | | | | | | | | | | | | | | | | | | | | | | | | | | | ✓ | | |
| Plastic bottle | ✓ | | | | | | | | | | | | | | | | | | | | | | | | | | | | | | ✓ | | |
| Tennis ball | | | | | | | | | | | ✓ | | ✓ | ✓ | | | | | | | | | | | | ✓ | | ✓ | | | | | ✓ |
| Credit card | | | | | | | | | ✓ | | | | | | | ✓ | | | | ✓ | | | | | | | | | | | | | |
| Scissors | | | | | | | | | | | | | | | | | | | ✓ | | | | | | | | | | | | | | |
| Screwdriver | | | | | ✓ | | | | | | | | | | | | | | | ✓ | | ✓ | | | | | | | | | | ✓ | |

**Table 2.** Selected objects with possible grasp patterns and minimum number of virtual fingers to execute the grasp.

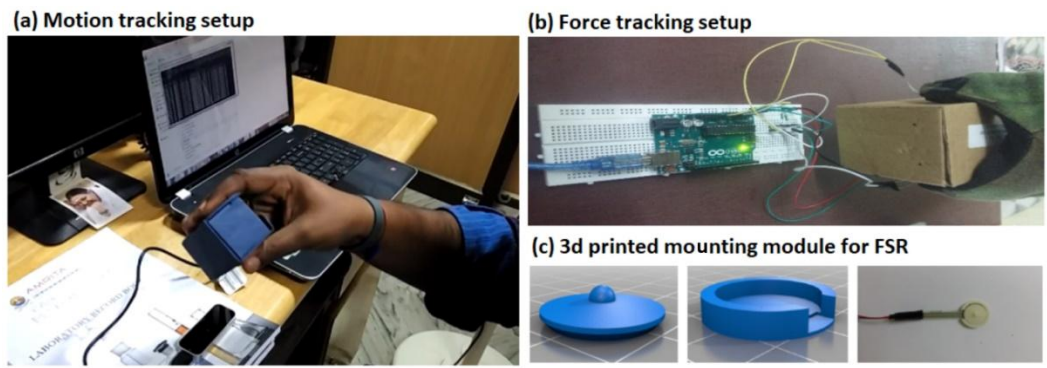

**Figure 2.** Experimental setup for characterization study: (a) Motion tracking setup, (b) Force tracking setup for hand fingers during grasping and (c) 3d printed mounting module for FSR.

The Leap Motion sensor is a small USB peripheral device which is placed on the table surface and connected to computer interface. Subjects grasped the 10 objects in the hemispherical workspace area of Leap motion sensor and traced the position and orientation parameters of user's hand with an average accuracy of 0.7 mm [63]. The experiment set up for tracking grasping forces in fingers as shown in Figure 2(b), users wore a glove with FSR to measure force and pressure. But the sensing ability of FSR is dependent on its contact area. To overcome this, a 3d printed mounting module placed on the sensing area of the FSR in the fingertips of gloves as shown in Figure 2(c). The values from FSR processed into multiple linear areas using an Arduino and measured the exerted forces in Newton(N). Experiment procedure in one trial consist of pick the object, grasp for 5 seconds, and place back the object and repeat for five trials for each object. In each trial subject's grasp parameters such as the position, speed, orientation, and force are tracked. Further analysis of these primary data leads to the modelling of multi-finger grasping haptics interface.

*4.3 Results and Discussions*

The 30 primary parameters were tracked and used for calculating the user's hand grasp movements and forces. Here the x-axis is defined as forward-backward, the y-axis as right-left, and the z-axis as up-down. Roll ($\gamma$) is taken to be about the x-axis, pitch ($\beta$) about the y-axis and yaw ($\alpha$) about the z-axis.

4.3.1. Direction, Trajectory, and Rotation of hand in grasping

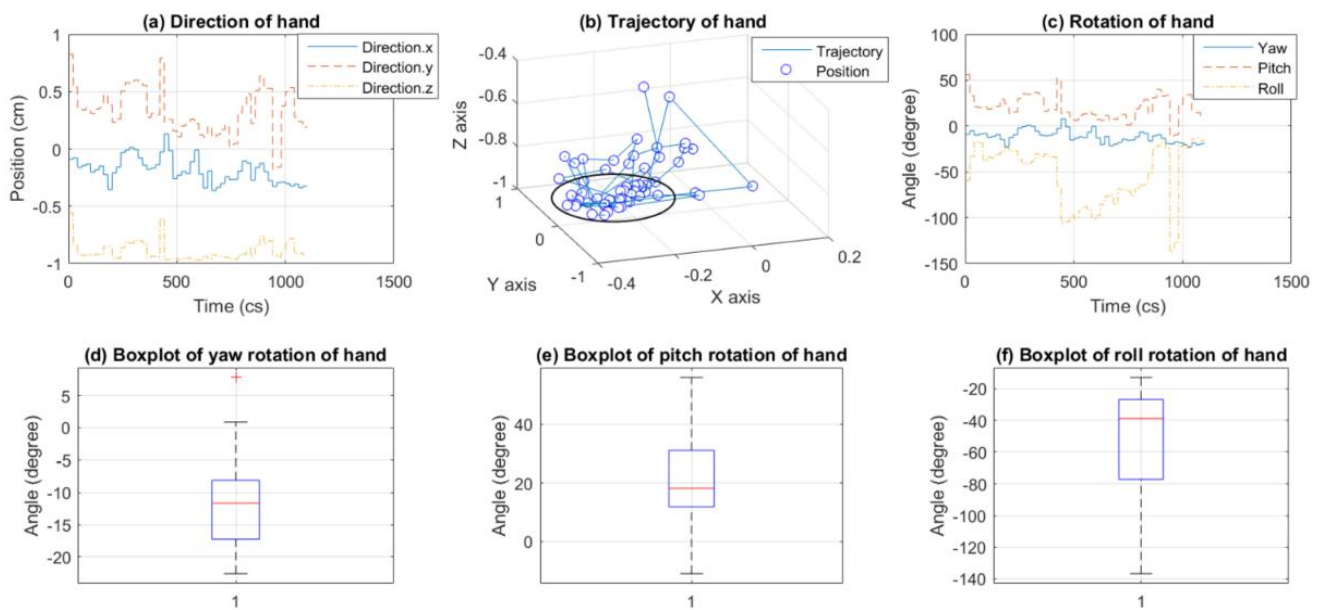

**Figure 3.** Plots for the movement of hand: (a) Direction of hand, (b) Trajectory of hand, (c) Rotation of hand, (d) Box plot of yaw rotation of the hand, (e) Box plot of pitch rotation, (f) Box plot of roll rotation of the hand.

The direction, trajectory, and rotation of hand in grasping (based on the experiment data set) is plotted as shown in Figure 3. The direction of the Hand in the 3D co-ordinate axes is shown in Figure 3(a). The x, y and z components of hand direction are spanning from -3 mm to 1 mm, -2 mm to 8 mm, and -10 mm to -6 mm respectively with standard deviation (SD) of 0.11, 0.20 and 0.08. The trajectory of hand movement direction is plotted in Figure 3(b). The movement in the y-direction is more than x and z-direction. The ellipse region in the plot represents the trajectory of hand exactly in the grasping time. Figure 3(c) plots the rotation of hand in the coordinate frame during grasping scenarios. The span of the roll is more than yaw and pitch. The boxplot of hand rotation clearly explains the distribution of the yaw, pitch, and roll of hand in grasping exercises shown in Figure 3(d)-(f). The minimum angle of hand yaw is -22.61° and the maximum is -0.86° with an average of -11.57° and SD of 6.7°. Inter Quartile Range (IQR) of hand yaw is 9.18°. The minimum angle of hand pitch is -11.06° and the maximum is 56.04° with an average of 20.04° and SD of 13.09°. Inter Quartile Range (IQR) of hand pitch is 19.21°. The minimum angle of hand roll is -136.69° and the maximum is -12.90° with an average of -53.68° and SD of 32.29°. Inter Quartile Range (IQR) of hand roll is 50.14°.

4.3.2. Position and trajectory of wrist and palm in grasping

The wrist is acted as the basement for most of the grasping scenarios. The information about the movement of the wrist is helpful for characterizing the basement of multi-finger gripper modules. The Position and trajectory of the wrist in grasping during the experiments were tracked and plotted as shown in Figure 4. Here the position of wrist n the 3D coordinate frame shows the span of position in x, y, and z-axes. It is clear from the plot that the span of the position of the wrist in grasping is average of 50 mm. Also, the plot of the trajectory of wrist shows a minimum workspace of movement for the wrist in grasping is in shape of a square with each side 50 mm. The thick ellipse region represents the data exactly during the grasping scenarios after noise filtering. Scatter plots in Figure 4 gives a clearer picture of the position of the wrist in the 3d space and 2D planes (x-y, y-z, and x-z).

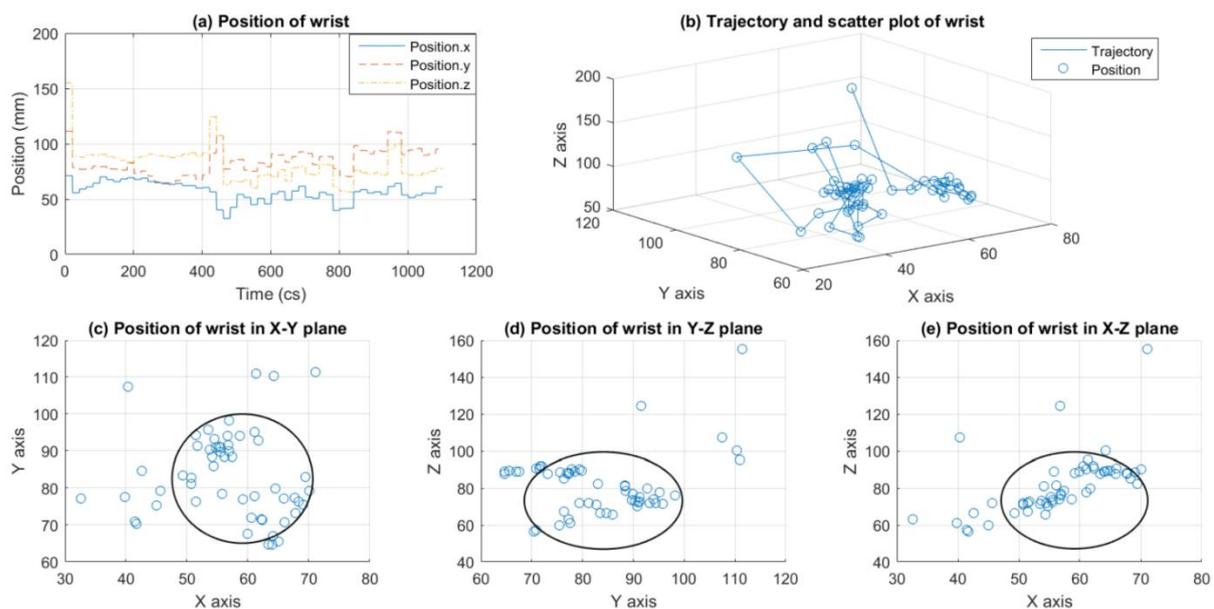

**Figure 4.** Plots for the movement of the wrist: (a) Position of the wrist, (b) Trajectory and scatter plot of wrist, (c) Position of the wrist in the x-y plane, (d) Position of the wrist in y-z plane, and (e) Position of the wrist in x-z plane.

Next, to the wrist, the palm is also an important part of better grasping. In some of the grasping types, the palm acts as an additional supportive area for the successful grasp. So, it is important to study the movement factors of palm while grasping. This will help to model the minimal number of virtual fingers (VF) for a general haptic grasping interface. The position and trajectory of the palm are plotted in Figure 5. It shows that there is not much deflection in the position of palm while grasping time in three axes. The trajectory of the palm shows the workspace of palm during the grasping scenarios. After filtering out the noise in the dataset, the maximum span in the 3D space for a grasping activity is 30 mm.

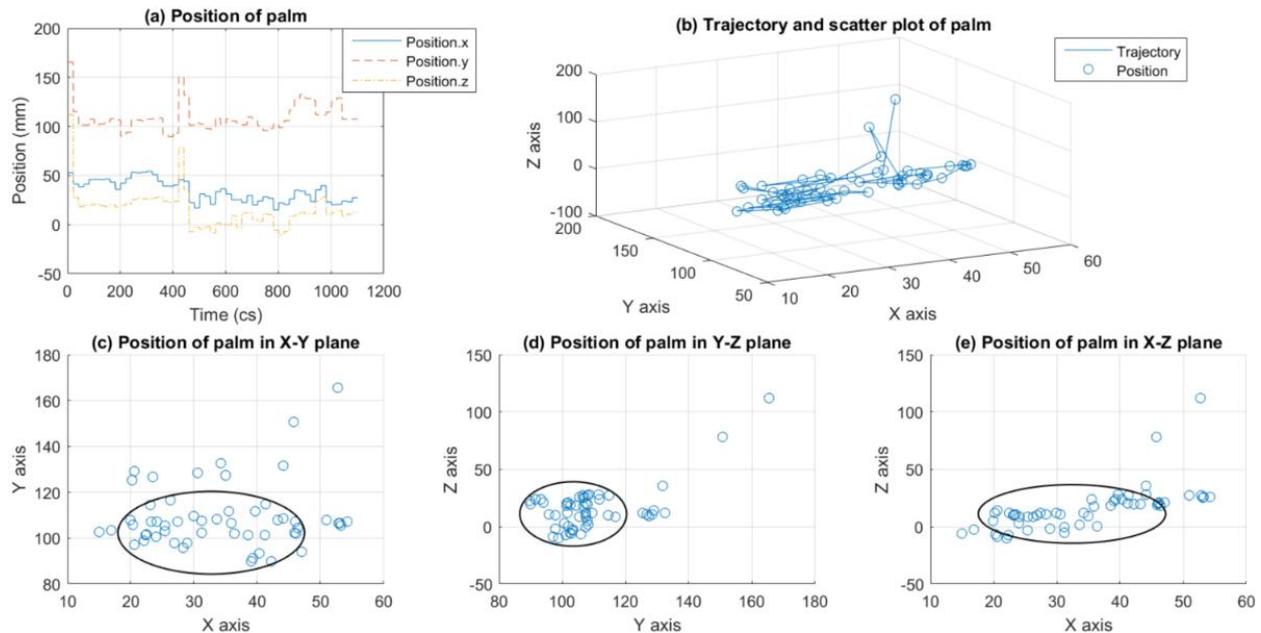

**Figure 5.** Position and trajectory of palm: (a) Position of palm, (b) Trajectory and scatter plot of palm, (c) Position of palm in the x-y plane, (d) Position of palm in the y-z plane, and (e) Position of palm in the x-z plane.

4.3.3. Angle of grasp

The angle of grasp is one of the common measurements used to describe grasping. Grasp angle describes the angle of the hand in grasping in relation to the Wrist position. The grasp angle influences comfort and easiness in grasping. A little bend in the wrist helps to maintain the grasp angle suitable for comfortable grasping of objects to get a proper grip. This will aid controlling the gripped objects easier. Figure 6 shows the various analysis plots of grasp angle tracked during the experimental scenarios.

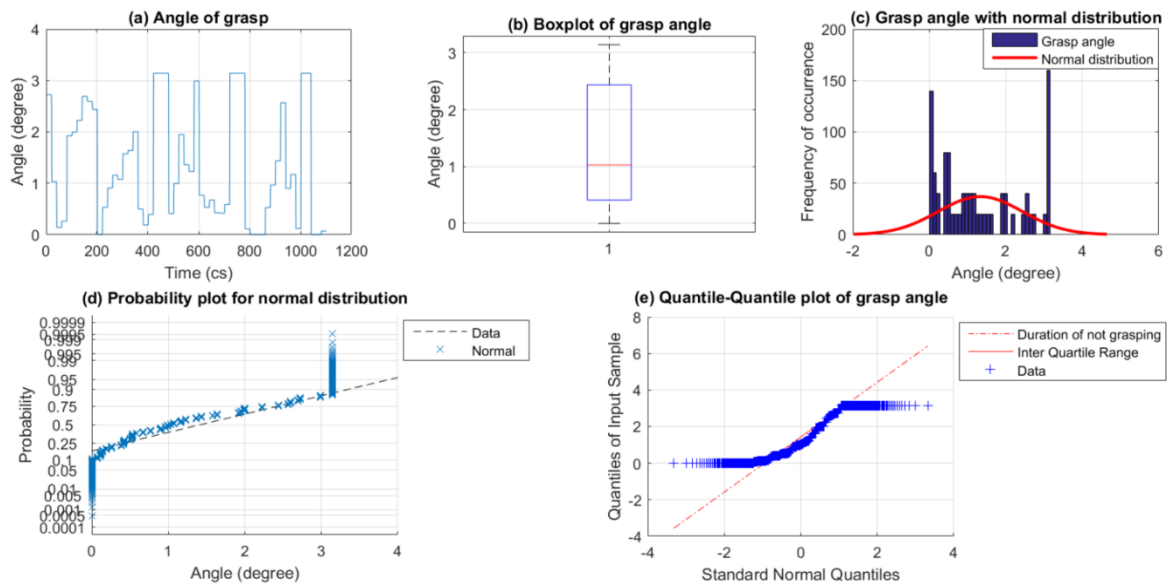

**Figure 6.** Plots for the angle of grasp: (a) Angle of grasp, (b) Box plot of grasp angle, (c) Grasp angle with a normal distribution, (d) Probability plot for normal distribution, and (e) Quantile-Quantile plot of grasp angle.

The angle of grasp is spanning from minimum 0° to maximum of 3.12° with an average value of 1.33° and SD of 1.10 as shown in Figure 6(a). The grasp angle data are charted as a box and whisker plot as shown in Figure 6(b). This will help to show the shape of the distribution, its central value, and its variability. The first quartile of the grasp angle values lies between 0° to 0.41°, second lies between 0.41° to 1.03°, third lies between 1.03° to 2.44° and the final lies between 2.44° to 3.14°. 75% of the grasp angle is below 2.44°. So, most of the grasp types can perform comfortably with a maximum angle of grasp 2.44°. The ideal maximum angle of grasp for the proposed gripper module should be between 2.5° to 4°.

Figure 6(c) plots a histogram of grasp angle in data using the number of bins equal to the square root of the number of elements in data and fits a normal density function. The bell curve fits the normal distribution with an SD of 1.10. 68% of the data falls within one SD of the mean 1.33°. The standard deviation controls the spread of the distribution. Here the larger standard deviation indicates that the data is spread out around the mean and the normal distribution is flat and wide. Figure 6(d) draws a normal probability plot, comparing the distribution of the grasp angle data to the normal distribution. The plot includes a reference line helped to judge whether the data follow a normal distribution. The plot shows that the normal line fit the data except the tails because of the outliers. Figure 6(e) displays a quantile-quantile plot of the sample quantiles of grasp angles versus theoretical quantiles from a normal distribution. The plot is close to linear in the IQR, so the distribution of grasp angle is normal during the grasping. In the duration of not grasping the plot shows the distribution is not normal.

4.3.4. Sphere of grasp

The spherical grip is the most used grasp in everyday life [64]. It is important to analyze the sphere of grasp in common grasping scenarios. Here the authors tracked the center and radius of the sphere of grasp in all the grasping scenarios in the experimental procedure. Figure 7 shows the various plots related to center and radius of the sphere of grasp. Through the experiment, the position of the center of the sphere is spanning from minimum (-147 mm, 48 mm, -48 mm) to maximum (181 mm, 259 mm, 85 mm) in x, y, and z-axes. After filtering out the non-grasping samples, the span of sphere grasp is reduced from minimum (-19 mm, 42 mm, -3 mm) to maximum (68 mm, 150 mm, 78 mm). The scattering and trajectory of the sphere of grasp shown in Figure 7(b).

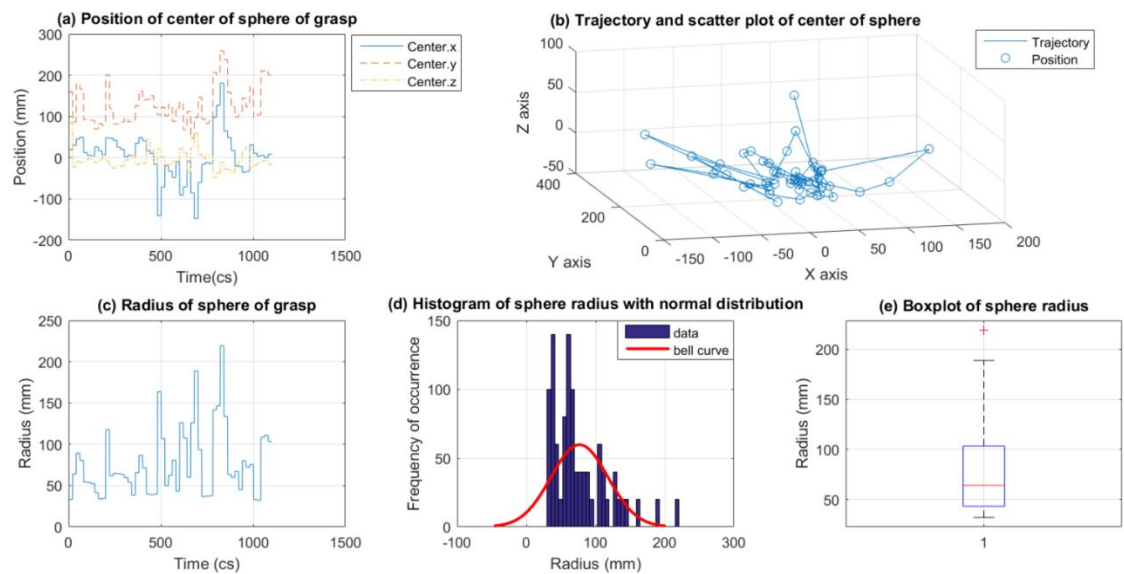

**Figure 7.** Center and radius plots for sphere of grasp: (a) Position of center of sphere of grasp, (b) Trajectory and scatter plot of center of sphere, (c) Radius of sphere of grasp, (d) Histogram of sphere radius with normal distribution, and (e) Box plot of sphere radius.

The volume of grasp can be represented by analyzing the radius of the sphere of grasp. Figure 7(c)-(e) shows the plots for the radius of the sphere of grasp. In the whole data tracked during the experiment, the radius of the sphere of grasp spanning from minimum 30 mm to 188 mm. In the box plot showing in Figure 7(e) 75% of the data is less than 10 mm and 50% of data is 64 mm. Again, after filtering out the noises the radius of the sphere of grasp is spanning from 31 mm to 50 mm. So, the authors targeting to model the grasping module with a radius of the sphere of grasp is 50 mm.

4.3.5. Distance of Pinch

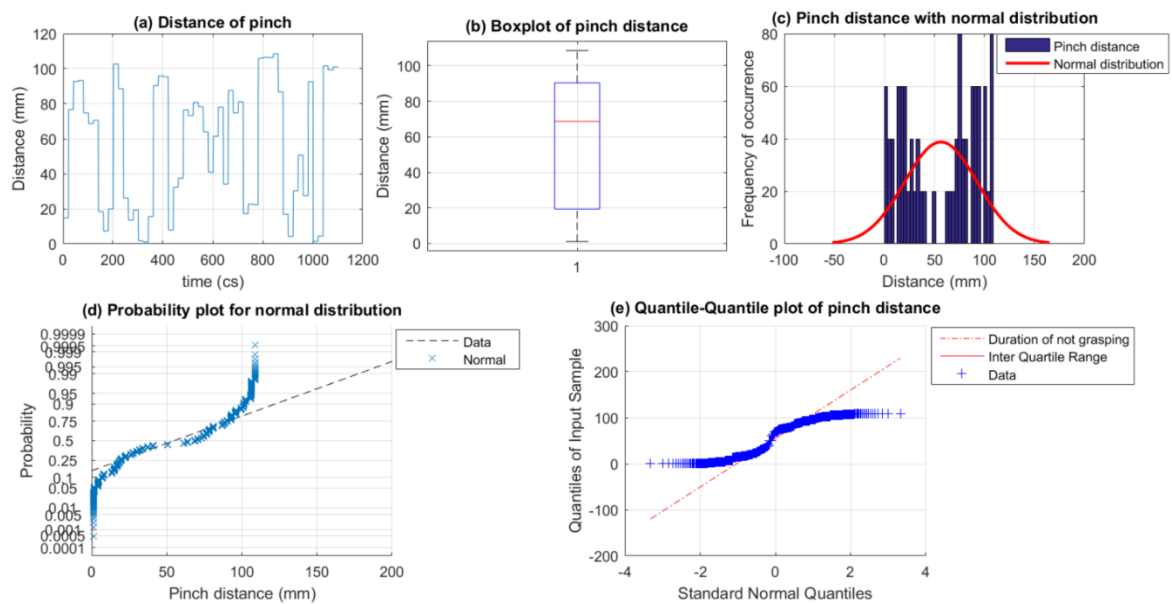

**Figure 8.** Plots for the distance of pinch: (a) Distance of pinch, (b) Box plot of pinch distance, (c) Pinch distance with a normal distribution, (d) Probability plot for normal distribution, and (e) Quantile-Quantile plot of pinch distance.

The pinch distance is the distance between two fingers in grasping. Pinch gestures are common for touch screens. The experiment interface tracked the pinch distance throughout the experiment and various plots are shown in Figure 8. The maximum pinch distance traced in the experiment setup is 110 mm. 75% of data is less than 90 mm and 50% is less than 70 mm as per the boxplot showed in Figure 8(b). The normal probability plot is shown in Figure 8(d) compared the distribution of the pinch distance to the normal distribution. The reference normal line fits the data in a range of 20 mm – 100 mm. Figure 8(e) displays a quantile-quantile plot of the sample quantiles of grasp angles versus theoretical quantiles from a normal distribution. This quantile-quantile plot is close to linear in the IQR, so the distribution of pinch distance is normal during the grasping. In the duration of not grasping the plot shows the distribution is not normal. So, the ideal values for the radius of the sphere of grasp for our proposed model are from 20 mm to 100 mm.

4.3.6. Finger motion parameters

When a person lifts any object, his fingers align in a particular way. This must be reflected by the device so that the user is at ease when he uses the device. The experiment setup traced the length and width of fingers and angular values between fingers in grasping objects differ in size and dimensions. These angular values measured between fingers must be replicated by the device. Unless the user feels comfortable when using the device, its intended purpose cannot be met. A base point was marked at the center of the outside of the palm. A line was drawn from the base point to the base of the middle finger. This was the baseline at 0 degrees as seen in Figure 9(a). Lines were also drawn to the base of the index and the thumb. The angles between the thumb and the index finger and the angles between the middle and the index fingers were measured during the grasping scenarios.

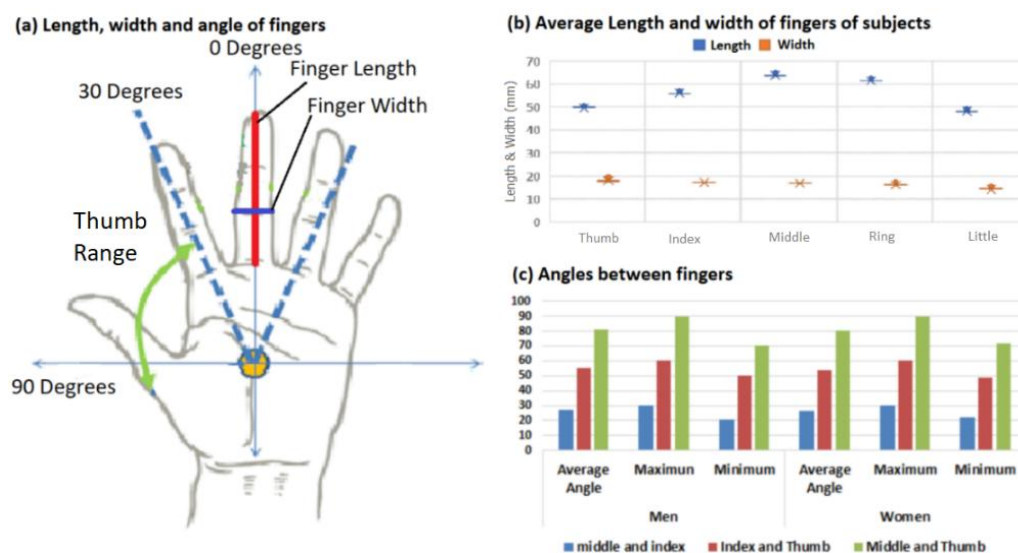

**Figure 9.** Length, width, and angle of fingers.

The average length and width of fingers of the subjects took part in the experiment shown in Figure 9(b). The average length and width of Thumb are 50 mm and 18 mm, the index finger is 57 mm and 17 mm, the middle finger is 67 mm and 17 mm, ring finger is 61 mm and 16 mm, and little finger is 48 mm and 14 mm. This will help the authors to model the wearable multi-finger grasping interface to fit the user's fingers. The measured angles were tabulated and plotted as shown in Figure 9(c), proves that angular measurements between fingers in grasping and lifting of objects does not depend on gender. The angle between thumb and index finger is in range of 50-60degree, index and middle finger is in range of 20-30 degree and thumb and middle finger is in range of 70-90 degree. This provided the data on how far apart the finger holders must be placed when designing the prototype for comfortable grasping and lifting of objects.

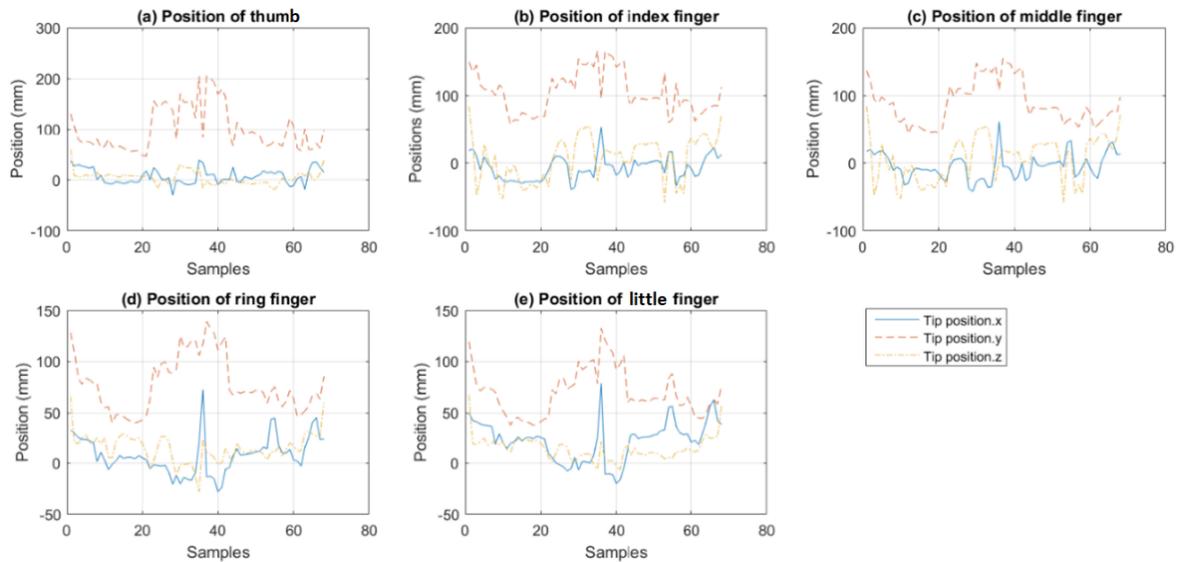

**Figure 10.** Position of movement of fingers

The motion parameters of fingers in grasping are so important to characterize and model the multi-finger grasping module. Here The authors traced all the five-finger's movement and force distributions. The movement of the tip of five fingers of the hand in all the grasping scenarios during the experiment was plotted in Figure 10. The movement of the fingertip is minimal in both x and z-axes is around -50 mm to +50 mm. The movement in x and z-axes are varied for fingers, especially it is very less in case of thumb, ring finger and little finger in the range of -20 mm to +20 mm. For index and middle finger, it is an almost same range of -50 mm to +50 mm. The movement in the y-axis is more compared to the other axes is around 50 mm to 200 mm for all fingers.

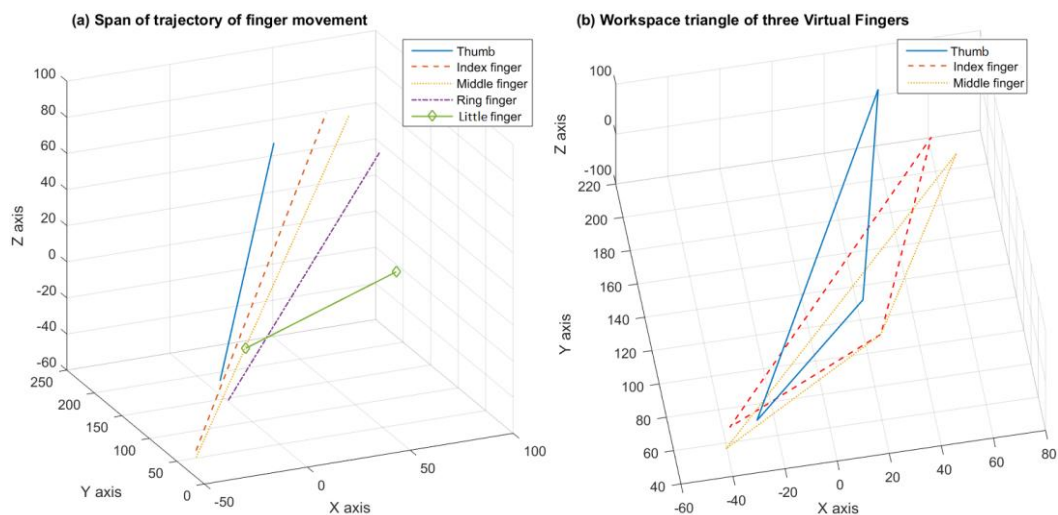

**Figure 11. (a)**. Span of the trajectory of fingers and (b) Workspace triangle of three Virtual Fingers.

The span of the trajectory of all five-finger movement in space during the grasping experiment is plotted together in Figure 11(a). This plot gave a clear idea about the active participation of all the fingers during grasping. Thumb is more active in grasping and gradually decreasing towards little finger. The thump, index and middle fingers more actively participate in the grasping scenarios than the ring and little fingers. Also, the workspace of the ring and little finger aligns with the middle finger as shown in Figure 11(a). So, authors proposed to group middle, ring, and little fingers to one Virtual Finger (VF) for multi-finger grasping interfaces. Thumb and index fingers can act as other two Virtual Fingers. The workspace triangle for three VF is shown in Figure 11(b).

4.3.7 Grasping forces on Fingers

The force values obtained through FSR sensors during the grasping exercises were tabulated and the average force was calculated for each person. As shown in Figure 12, forces are not more than 10N was experienced by the user. It can also be seen that the thumb experienced the maximum force, whereas the middle and the index finger experienced somewhat similar forces.

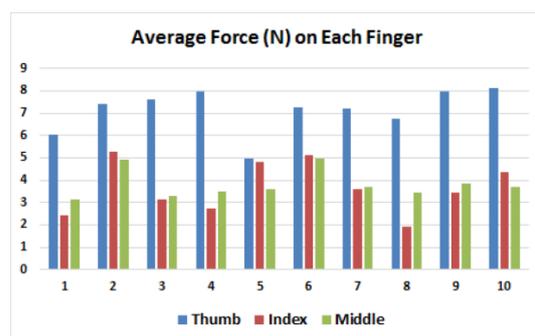

**Figure 12.** Average grasping force values on each finger.

## 5. Computational Model for Tripod Haptic Grasp

The multi-finger perception studies confirmed the hypothesis that the minimal configuration that allows most of the grasps is the three-finger tripod grasp. Based on the human grasp analysis, a conceptual, computational model is presented here for the three-finger tripod haptic grasping interface. Previous researchers worked on force models [65] and virtual linkages [66] for multi-grasp manipulations.

The concept of the virtual finger [15] has been postulated as an abstract representation through which the human brain plans are grasping tasks [60]. The virtual finger is a functional unit of several fingers work together comprised of at least one real physical finger (which may include the palm). This effectively reduces the many degrees of the human hand to those that are deemed necessary to perform the grasping task. This concept replaces the analysis of the mechanical degrees of freedom of individual fingers by the analysis of the functional roles of forces being applied in a grasp. Here the concept of the virtual finger was implemented to reduce the realistic five-finger grasping to virtual three-finger grasping. The characterization study revised the existing virtual fingers allocation and replaced with new tripod virtual fingers allocation.

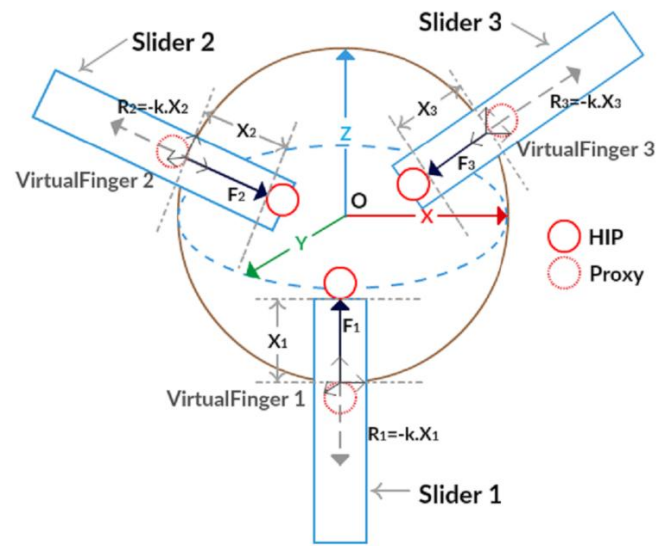

**Figure 13.** Haptic computation model for three-finger tripod haptic grasping interface.

The haptic computation model is shown in Figure 13 was implemented to create the suitable force feedback for the tripod haptic grasping interface. Each finger holder is connected to each slider and these sliders are responsible for creating forces to each finger attachment point on the grasping interface through the closed-loop belt system. Three proxies and Haptic Interactive Points (HIP) are assigned for three fingers. Based on the position information received from each slider, the position of proxy and HIP are updated. Collision detection algorithms detected the collision of each fingertip with the virtual object, these collision points act as simple virtual walls. Then applying the God object algorithms to these three proxies and calculated the resultant forces as shown in Figure 13. Each actuator in the slider generates forces and these forces are transferred to the fingertip through the attached finger holder. In this proposed model for multi-finger tripod haptic grasping, the thumb is assigned as $VF_1$, Index finger as $VF_2$ and other three fingers as a single virtual finger $VF_3$. The proposed model including object and virtual fingers in the virtual interface measures collision and progress of interactions in terms of applying forces and movements of the fingers and computes force feedback to be provided by the haptic interfaces. It covers the dependence of the force feedback and the effect of finger motions on the tripod grasp. A basic grasping touch in virtual reality is defined as a touch that provides vertical force feedbacks on the contact surface coincident with the reverse direction of the fingers in the space.

Forces and moments from the individual virtual fingers are considered for the rendering of resultant forces for the model. Three proxies and Haptic Interactive Points (HIP) are assigned for three virtual fingers. Based on the position information received from each $VF_i$, the position of proxy and HIP are updated. When collision detection algorithms detected the collision of each $VF_i$ with the virtual object, these collision points act as simple virtual walls. The authors assume that only forces act at the grasp points. Let $u_{ij}$ is the unit vector along the virtual finger *i* to *j*, $v_i$ be the vector from the object reference point *o* to the virtual finger grasp point. Let $F_i$ be the force exerted on objects through each virtual finger $VF_i$ and $F$ be the vector. Based on the object shape, size, and stiffness the applying forces on objects by the virtual fingers are different.

$$F_i = k\, X_i \tag{1}$$

where *i* = 1,2,3.

$$F = [F_1 \quad F_2 \quad F_3]^T \tag{2}$$

Grasp perception rate depends on the force applied by the fingers $F_i$, the area of contact between the fingers and the object A, the distance between the proxy and HIP $X$ and the change in position of the HIP $\Delta X$. Mathematically, the perception rate is

$$P = \frac{k\,\Delta X\,F_i}{A} \tag{3}$$

where $k$ is a constant that depends on the material of the object. Let $R_i$ be the resultant force exerted on each virtual finger $VF_i$ and $R$ be the vector. Then applying the God object algorithms [67] to three proxies and calculated the resultant forces $R_i$.

$$R = [R_1 \quad R_2 \quad R_3]^T \tag{4}$$

The relationship between the applied forces $F$ and resultant forces $R$ would be given by

$$R = uF \tag{5}$$

were

$$u = \begin{bmatrix} u_{11} & u_{12} & u_{13} \\ u_{21} & u_{22} & u_{23} \\ u_{31} & u_{32} & u_{33} \end{bmatrix} \tag{6}$$

The relationships between the resultant force $R$, the resultant moment $m$, and the applied forces $F$ are given by

$$F = v\,[R \quad m]^T \tag{7}$$

were

$$v = [v_1 \quad v_2 \quad v_3]^T \tag{8}$$

and

$$m = [m_1 \quad m_2 \quad m_3]^T \tag{9}$$

Force feedback on the grasping scenarios along virtual fingers has two components; the force feedback resisting the grasping motion and a frictional force ($F_r$), represented by

$$R = uF + F_r \tag{10}$$

An experiment was done to evaluate the rendered force to the fingers through the virtual fingers using the proposed tripod haptic grasp model. The results plotted as shown in Figure 14. Initially, the grasping forces in the real grasping cases were tracked using the FSR and plotted as shown in Figure 14(a). In case of the real grasping scenarios, the forces exerted by the thumb is in the range of 6-8N which is greater than the force exerted by the index and middle finger. The force exerted by the index and middle fingers in in the range of 2-5N and 3-5N respectively. The same experiment was carried out to measure the rendered forces in haptic grasping interfaces during virtual grasping. Force is measured using the FSR sensor. The FSR was placed inside the finger holder of the gripper at the location of the fingertip contact. The sensing part of FSR facing the flat area of the holder and the user keeps a finger on top of the FSR. The motors were actuated by the haptic rendering model. The tracked rendered forces were plotted in Figure 14(b). This graph shows that the range of rendered forces in virtual grasping is almost in the same range of forces in real grasping scenarios. The range of force rendered at the Thumb, index and middle finger holders is 5.7-7.5N, 2.6-4.4N and 3-4.5N respectively. This demonstrates that the three-finger haptic grasping interfaces able to provide realistic grasping haptic feedbacks to the users. Also, the experiment setup was traced the peak force that can provide the gripper device. As plotted in Figure 14(c), the device attained a peak force of 10.4N, 10.1N, and 10.2N at Thumb, index, and middle finger holders respectively. It was found that the interface can give approximately 10N force which is greater than the force when lifting objects as found out in Section 3.1. Finally, the Fig-ure 14(d) shows the performance index of three-finger haptic grasping interface in haptic feedback and finger manipulability.

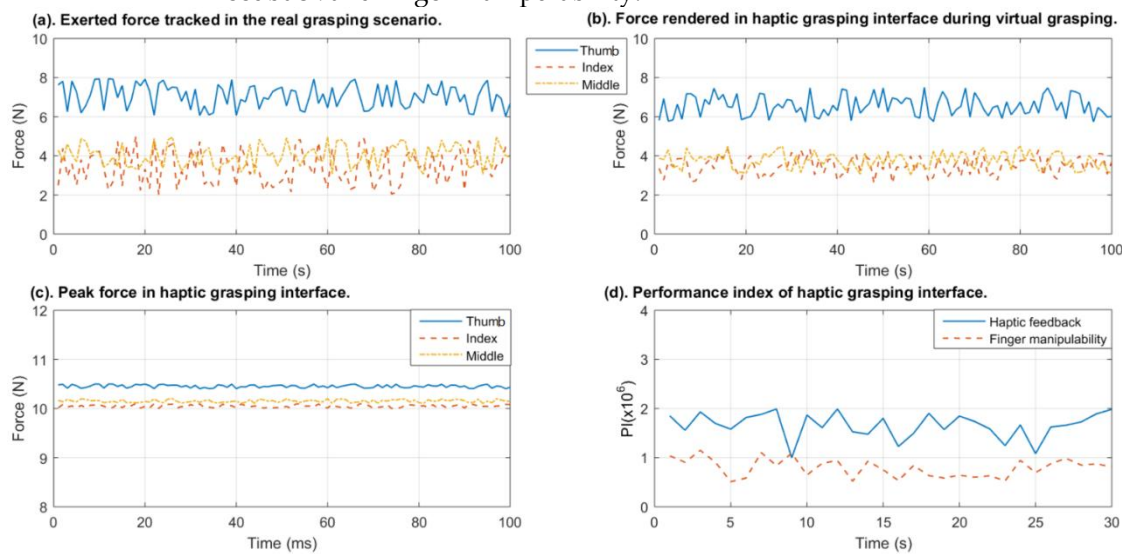

**Figure 14.** Evaluation results of the haptic grasping interface: (a) Exerted force tracked in the real grasping scenarios, (b) Force rendered in haptic grasping interface during virtual grasping, (c) Peak force generated in the haptic grasping interface, and (d) Performance index of the haptic grasping interface.

## 6. Conclusions

The aim of this research was to come up with an analysis, model, and design of a three-finger haptic interface. A detailed literature review was carried out regards the anatomy, kinematics, and dynamics of hand. Also, a focused survey on the different human grasps, prehension patterns and grasp taxonomy. As part of this work, authors carried out characterization studies in most of the major aspects of human grasping. The position, orientation, and forces of hand, wrist, palm, and fingers were analyzed, and the results were discussed. The characterization studies confirmed the hypothesis that the minimal configuration that allows most of the grasps in grasp taxonomy is the three fingers grasping. This detailed characterization study leads to the design of a three-finger haptic grasping interface as an extension.

As a future work, I am planning to work on the multi-Finger grasping interface for bimanual scenarios which provides the users a complete immersed grasping manipulation in the virtual and remote environment.

**Funding:** This research received no external funding.

**Institutional Review Board Statement:** Not applicable.

**Informed Consent Statement: Not applicable.**

**Data Availability Statement: Not applicable.**

**Acknowledgments:** I wholeheartedly thank the wonderful team at AMMACHI Labs for their support, encouragement and for providing constructive criticism and valuable inputs during the various stages of this work.

**Conflicts of Interest:** The author declares no conflicts of interest. The funders had no role in the design of the study; in the collection, analyses, or interpretation of data; in the writing of the manuscript; or in the decision to publish the results.